# News Article Retrieval in Context for Event-centric Narrative Creation

Nikos Voskarides*
Amazon
Barcelona, Spain
nvvoskar@amazon.com

Edgar Meij
Bloomberg
London, United Kingdom
emeij@bloomberg.net

Sabrina Sauer
University of Groningen
Groningen, The Netherlands
s.c.sauer@rug.nl

Maarten de Rijke
University of Amsterdam & Ahold Delhaize Research
Amsterdam, The Netherlands
m.derijke@uva.nl

## ABSTRACT

Writers such as journalists often use automatic tools to find relevant content to include in their narratives. In this paper, we focus on supporting writers in the news domain to develop event-centric narratives. Given an incomplete narrative that specifies a main event and a context, we aim to retrieve news articles that discuss relevant events that would enable the continuation of the narrative. We formally define this task and propose a retrieval dataset construction procedure that relies on existing news articles to simulate incomplete narratives and relevant articles. Experiments on two datasets derived from this procedure show that state-of-the-art lexical and semantic rankers are not sufficient for this task. We show that combining those with a ranker that ranks articles by reverse chronological order outperforms those rankers alone. We also perform an in-depth quantitative and qualitative analysis of the results that sheds light on the characteristics of this task.

## 1 INTRODUCTION

Professional writers such as journalists generate narratives centered around specific events or topics. As shown in recent studies, such writers envision automatic systems that suggest material relevant to the narrative they are creating [10, 19]. This material may provide background information or connections that can help writers generate new angles on the narrative and thus help engage the reader [24].

Previous work has focused on developing automatic systems to support writers explore content relevant to the narrative they are writing about. Such systems use content originating from various sources such as as social media [8, 11, 53], political speeches and conference transcripts [29], or news articles [30].

*Research conducted when the author was at the University of Amsterdam.

Writers in the news domain often develop narratives around a single main event, and refer to other, related events that can serve different functions in relation to the narrative [44]. These include explaining the cause or the context of the main event or providing supporting information [4]. Recent work has focused on automatically profiling news article content (i.e., paragraphs or sentences) in relation to their discourse function [4, 52].

In this paper, instead of profiling existing narratives, we consider a scenario where a writer has generated an incomplete narrative about a specific event up to a certain point, and aims to explore other news articles that discuss relevant events to include in their narrative. A news article that discusses a different event from the past is relevant to the writer's incomplete narrative if it relates to the narrative's main event and to the *narrative's context.* Relevance to the narrative's main event is topical in nature but, importantly, relevance to the narrative's context is not only topical: to be relevant to the narrative's context, a news article should enable the continuation of the narrative by expanding the narrative discourse [2]. Table 1 shows an example of an incomplete narrative and a news article relevant to it. The relevant article discusses an event about a subject mentioned in the narrative context (*Italy*). Here, the relevant news article is relevant to the topic of the incomplete narrative (*migration crisis*) and also relevant to the narrative context in the sense that it is used by the writer to expand the narrative by making a comparison: the previous government of Italy was more welcoming to immigrants than the current. To avoid confusion, in the remainder of this paper *relevance* without further restriction or scope is taken to mean both *topical relevance* and *relevance to the narrative context.*

We model the problem of finding a relevant news article given an incomplete narrative as a retrieval task where the query is an incomplete narrative and the unit of retrieval is a news article. We automatically generate retrieval datasets for this task by harvesting links from existing narratives manually created by journalists. Using the generated datasets, we analyze the characteristics of this task and study the performance of different rankers on this task. We find that state-of-the-art lexical and semantic rankers are not sufficient for this task and that combining those with a ranker that ranks articles by their reverse chronological order outperforms those rankers alone.

Our main contributions are: (i) we propose the task of news article retrieval in context for event-centric narrative creation; (ii) we propose an automatic retrieval dataset construction procedure for this task; and (iii) we empirically evaluate the performance of different rankers on this task and perform an in-depth analysis of the results to better understand the characteristics of this task.

**Table 1: Example incomplete narrative $q$ (consisting of a main event $e$ and a narrative context $c$), and a news article $d^*$ that is relevant to $q$ because it is relevant to both the main event $e$ and to the narrative context $c$ in the sense explained in the main text.**

| |
|---|
| **Incomplete narrative** $q$<br>**– Main event** ($e$)<br>(#1) Malta's armed forces storm merchant ship taken over by rescued migrants.<br>(#2) Maltese armed forces on Thursday stormed a merchant vessel taken over by rescued migrants who were allegedly demanding to be transported to Europe, rather than back to Libya.<br>**– Narrative context** ($c$)<br>(#3) In earlier years of Europe's migration crisis—when flows from the Middle East and North Africa were much higher—the Mediterranean was patrolled by Italian and European vessels, as well as by humanitarian groups, which would rescue migrants from flimsy dinghies and transport them to safety, typically to Italy. |
| **Relevant news article** ($d^*$)<br>(#4) Italy's new government sends immigration message by rejecting rescue ship<br>(#5) Italy's new populist government has delivered a jolt to European migration politics, prompting a diplomatic standoff with its refusal to accept a rescue vessel overloaded with migrants. |

# 2 PROBLEM STATEMENT

## 2.1 Preliminaries

A *news article* $d$ published at time $t$ consists of its headline $H$—which introduces the topic of the article [44]—and a sequence of paragraphs $p_1, p_2, \ldots$. Each paragraph $p_i$ consists of a sequence of sentences $a_{i,1}, a_{i,2}, \ldots$.

The *lead paragraph* $L$ of a news article $d$ is its first paragraph $p_1$, which usually summarizes the main topic of the article [44].

An *event* $e$ is characterized by interactions between entities such as countries, organizations, or individuals—that deviate from typical interaction patterns [3]. We assume that each news article $d$ is associated with a single main event $e$, and that $e$ is described by the headline $H$ and the lead paragraph $L$ of $d$ [4].

A *link sentence* $a_{i,j}$ in article $d$ is a sentence that contains a hyperlink to a news article $d^*$.

A *context* is a sequence of sentences already generated by the writer that introduces a new idea or subtopic in a narrative.

A *query* $q = (e, c, t)$ is an incomplete narrative at time $t$ that consists of an event $e$ and a context $c$.

## 2.2 Task definition

The task of *news article retrieval in context for event-centric narrative creation* is defined as follows. Given a query $q = (e, c, t)$ and a collection of news articles $D$ published before time $t$, we need to rank articles in $D$ w.r.t. their relevance to $q = (e, c, t)$. Importantly, "relevance to $e$" is to be interpreted as topical, whereas "relevance to $c$" is not only topical, but it should also enable the continuation of the narrative by expanding the narrative discourse [2]. "Relevance to $q$" is taken to mean the same as "relevance to $e$ and to $c$". An article relevant to $q$ can thus be used by the writer to create the next sentence in the yet incomplete narrative. Table 1 shows an example query $q$ and a relevant news article $d^*$ published at time $t^* < t$.

# 3 RETRIEVAL DATASET CONSTRUCTION

## 3.1 Dataset construction procedure

In order to construct a retrieval dataset for our news article retrieval task, we rely on existing news articles to simulate incomplete narratives as well as relevant documents. We capitalize on the fact that (complete) news articles often contain links to other news articles manually inserted by journalists in the form of hyperlinks.

The automatic retrieval dataset construction procedure that we propose takes as input a news article $d$ and outputs a set of $(q, d^*)$ pairs, where $q = (e, c, t)$ is a query and $d^*$ is the (unique) relevant news article to $q$. Note that $e$ is described by the headline $H$ and the lead paragraph $L$ of $d$ (see Section 2.1).

In order to construct the context $c$ of $q$, we iteratively look for link sentences $a_{i,j}$ in $d$ that contain a hyperlink to another news article $d^*$. We enforce $i > 1$ so that the paragraph where the link sentence appears is after the lead paragraph. We also enforce $j > 1$ motivated by the fact that links after the first sentence of a paragraph are tightly related to the main idea of the paragraph, therefore the sentences preceding the link sentence can be considered as context [16]. If such a link sentence $a_{i,j}$ exists, we consider the sentences $a_{i,1}, \ldots, a_{i,j-1}$ as the narrative context $c$ and the article $d^*$ as the relevant article for $q$.

*Example.* To illustrate the procedure described above, consider the example in Table 1. Sentences #1 and #2 in Table 1 are the headline and lead paragraph of a news article $d$, respectively. Sentence #3 in Table 1 is the first sentence $a_{i,j-1}$ of a paragraph $p_i$, $i > 1$ in $d$, which constitutes the narrative context $c$. The link sentence $a_{i,j}$ (not shown in the table) is:

> But over the past year, *Italy has closed its ports* to migrants rescued by humanitarian boats.

where the part in italics is (the anchor text of) a hyperlink to the relevant news article $d^*$ shown in Table 1, where sentences #4 and #5 are the headline and lead of $d^*$, respectively.

## 3.2 Retrieval dataset description

We consider two collections of news articles written in English and published by major newspapers. The first is a set of news articles published by The Washington Post (WaPo), released by the TREC News Track [41]. It contains 671,947 news articles and blog posts published from January 2012 to December 2019. The second is a set of news articles published by The Guardian, between November 2013 to June 2017, which we have crawled ourselves. We have also crawled the out-links of each article in this set; the final set contains

**Table 2: Statistics of the retrieval datasets derived from the WaPo and Guardian newspaper collections. Because of the way we construct the retrieval datasets (see Section 3.1), each query has a single relevant news article.**

| Dataset | Split | $\#q$ | #uniq. $d$ | #uniq. $d^*$ | Link sentence ($a_{i,j}$) $i$ mean/ median | $j$ mean/ median |
|---|---|---|---|---|---|---|
| WaPo | Train | 32,963 | 23,537 | 24,279 | 7.9/7 | 2.5/2 |
| | Dev. | 1,831 | 1,286 | 1,585 | 8.4/8 | 2.4/2 |
| | Test | 1,832 | 1,216 | 1,555 | 9.1/9 | 2.4/2 |
| Guardian | Train | 31,329 | 21,730 | 22,935 | 7.3/6 | 2.4/2 |
| | Dev. | 1,740 | 1,128 | 1,526 | 8.0/7 | 2.4/2 |
| | Test | 1,742 | 1,064 | 1,532 | 7.3/7 | 2.5/2 |

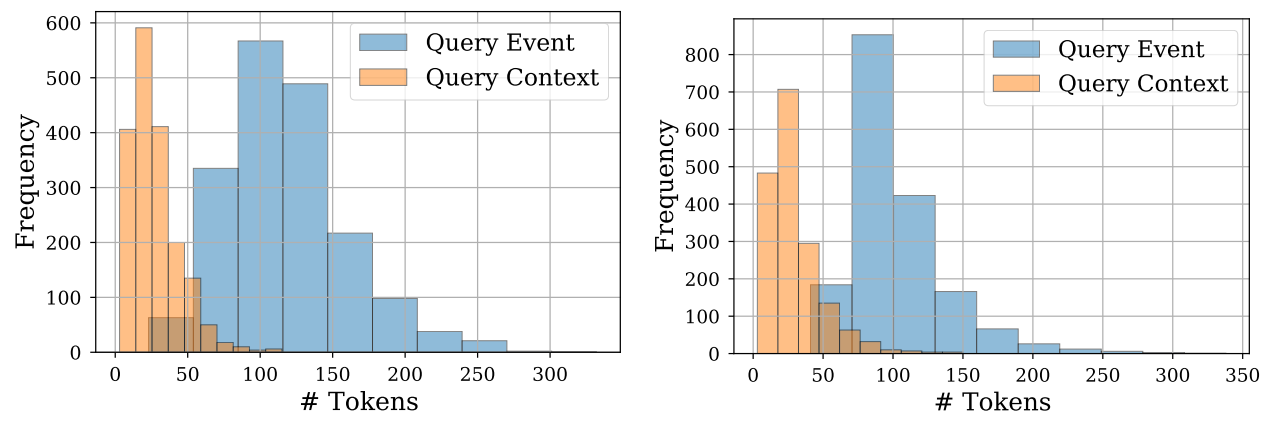

**(1) WaPo dev.** **(2) Guardian dev.**

**Figure 1: Histogram of the number of tokens in the query event $e$ and the query context $e$.**

572,639 news articles published between January 2000 and March 2018.

The articles in both newspapers cover multiple genres and domains. In order to ensure that the news articles describe real-world events, we filter out blog posts and opinion news articles, and only keep articles in the following domains: *news*, *world*, *business*, *environment*, *technology*, *society*, *science*, *culture*, *education*, *global*, *healthcare*, *media*, *money*, *teacher*, *local*, *national*. After filtering for genre and domain, we are left with 386,196 articles in WaPo and 185,034 in The Guardian.

We then apply the dataset construction procedure described in Section 3.1 to construct a retrieval dataset for both collections. We split the retrieval datasets chronologically and keep the first 90% for training, the next 5% for development, and the last 5% for testing. Table 2 shows basic statistics for both retrieval datasets. Figure 1 shows a histogram of the number of tokens in the query event $e$ and the query context $c$. We observe that the query context is shorter than the query event in both datasets. Also, the query event is longer in WaPo than in Guardian because the way those newspapers perform paragraph splitting is different.

Figure 2 shows a histogram of the difference in number of days between the publication date of the query and the publication date of the relevant news article on the development sets of the two datasets. The retrieval datasets have a strong recency bias, which is in line with studies on content generation in the news domain [32]. Typical examples of recent, relevant articles are those discussing a previous development of a query event or of an event mentioned in the narrative context. And a typical example of a less recent, relevant article can be found when discussing an event that

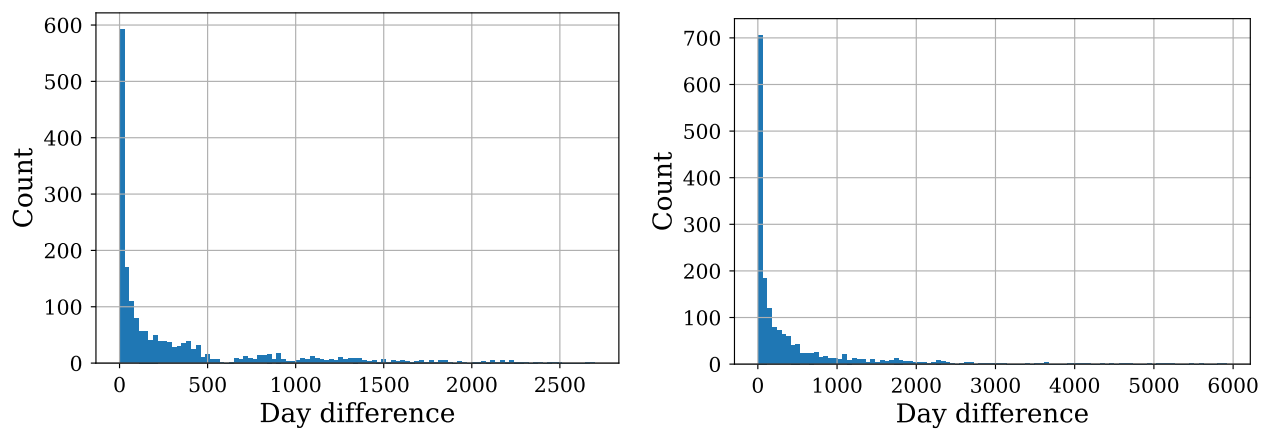

**(1) WaPo dev.** **(2) Guardian dev.**

**Figure 2: Histogram of the difference in the number of days between the publication date of the query and of its relevant news article.**

**Table 3: Results of the annotation exercise: assessing relevance of document $d^*$ w.r.t. $e$ only (Task 1), and then $c$ (Task 2). We show the fraction of times the annotator labeled a sample as positive for the task.**

| Dataset | Task 1 | Task 2 | Either |
|---|---|---|---|
| WaPo | 0.90 | 0.77 | 0.91 |
| Guardian | 0.85 | 0.83 | 0.92 |

is similar to one mentioned in the query (e.g., an earthquake) but involving different entities (e.g., a person, location, or organization).

### 3.3 Retrieval dataset quality

The dataset construction procedure we described in Section 3.1 assumes that an article $d^*$ is relevant to $q$ because the writer has chosen to link to it in a particular context, which is a fair assumption to make. Nevertheless, we further assess the quality of the automatically constructed retrieval datasets with respect to our task definition (Section 2.2) by performing two annotation tasks. In the first task, we show $e$ and $d^*$ to a human annotator and ask whether they understand their connection (binary). In the second task, which is done after the completion of the first task, we additionally show the context $c$ and ask whether it enhances their understanding of the connection of $e$ and $d^*$ (binary). The two tasks can help us validate whether $d^*$ is topically relevant to $e$, and relevant to $c$ in a way that enables the continuation of the narrative (Section 2.2).

One annotator annotated 100 examples from the development set of each dataset (i.e., 200 examples in total). In order to assess the quality of the annotations, a second assessor annotated a subset of 50 examples from each dataset (100 examples in total). The Cohen's $\kappa$ [6] score is 0.61 for Task 1 (substantial agreement) and 0.50 for Task 2 (moderate agreement).

The results can be seen in Table 3. We see that, for both datasets, the context $c$ enhances the understanding of the connection to $d^*$ for more than 3/4 of the cases (Task 2). Also, for the vast majority of the cases, either the event $e$ or the context $c$ is sufficient to understand the connection (third column). We conclude that the automatic dataset construction procedure we proposed in Section 3.1 can produce reliable datasets for the task of news article retrieval in context for event-centric narrative creation.

# 4 RETRIEVAL METHOD

We follow a standard two-step retrieval pipeline that consists of (i) an unsupervised initial retrieval step, and (ii) a re-ranking step [48]. Note that we do not focus on proposing new methods but rather on studying existing ones on this novel task.

## 4.1 Initial retrieval

In this step, we score each news article $d$ in $D$ w.r.t. $q = (e, c, t)$ to obtain the initial ranked list $L_1$. Here, we are interested in achieving high recall at lower depths in the ranking, since this step is followed by a more sophisticated reranking step. We use BM25 [36], an unsupervised lexical matching function, which is effective for ad-hoc retrieval and other tasks, such as question answering [50]. In order to construct the lexical query, we simply concatenate $e$ and $c$.

## 4.2 Reranking

Here we rerank the initial ranked list $L_1$ obtained in the previous step by combining the results of multiple rankers using Reciprocal Rank Fusion (RRF) [7], an unsupervised ranking fusion function that is effective in combining state-of-the-art rankers [27, 46]. RRF is defined as follows:

$$\sum_{L \in L'} \frac{1}{k + rank(d, L)}, \quad (1)$$

where $L'$ is a set of ranked lists, $rank(d, L)$ is the rank of article $d$ in the ranked list $L$, and $k$ is a parameter, set to its default value (60).

We use the following rankers:

**BM25** The initial retrieval step ranker (Section 4.1), often used in combination with more sophisticated ranking models [28].

**BERT** BERT [9] has recently achieved state-of-the-art performance for retrieval and recommendation tasks in the news domain [49, 51]. BERT has been shown to prefer semantic matches and it is often used in combination with lexical matching ranking functions [35]. Given the query $q$ and a candidate news article $d$, we follow [29] and construct the input to BERT as follows: [<CLS> $e$ <unused> $c$ <SEP> $d$], where <CLS> is a special token, <unused> is a special token that informs the model where the context begins and <SEP> is a special token that informs the model where the document $d$ begins. We add a dropout layer on top of the <CLS> token, and a linear layer with a scalar output to obtain the final matching score, which is used to rank the articles in $L_1$. Note that, because of the limit of BERT in the number of tokens, we only take into account the headline and lead of $d$.

**Recency** This ranker simply sorts the candidate articles in $L_1$ by their reversed chronological order.

Note that we have also experimented with using the scores of the above rankers as features in supervised learning to rank models but they only gave minor improvements over RRF. Thus we do not discuss them in this paper.

# 5 EXPERIMENTAL SETUP

## 5.1 Evaluation metrics

We use standard IR metrics: Mean Reciprocal Rank (MRR) and recall at different cut-offs (R@20, R@1000). Because of the way we construct our dataset (Section 3.1), we only have one relevant news article per query and thus MRR is equivalent to MAP. We use a cut-off of 20 at recall since we expect writers to be willing to navigate the ranked list to lower positions [23, 38, 45]. We report on statistical significance with a paired two-tailed t-test.

**Table 4: Initial retrieval performance of BM25 on the test sets for different variations of the query $q = (e, c, t)$, or the link sentence (LS).**

| | WaPo | | Guardian | |
|---|---|---|---|---|
| **Query** | MRR | R@1000 | MRR | R@1000 |
| $e$ | 0.117 | 0.745 | 0.104 | 0.723 |
| $c$ | 0.167 | 0.737 | **0.154** | 0.714 |
| $e$ & $c$ | **0.172** | **0.832** | 0.149 | **0.806** |
| LS | 0.459 | 0.944 | 0.427 | 0.929 |

## 5.2 Implementation and hyperparameters

We use the BM25 implementation of Anserini [50] with default parameters and retrieve the top-1000 articles (Section 4.1).

We use the OpenNIR implementation of BERT for retrieval [28]. We fine-tune the *bert-base* pre-trained model on the training set of each of our datasets separately. We assign a maximum 300 tokens for the query $q$ and 200 for the article $d$. We use a batch size of 16 with gradient accumulation of 2; we apply max grad norm of 1, and tune the following hyperparameters for MRR on the development set of each dataset separately: number of negatives $\{1, 2, 3\}$ and learning rate $\{5e-6, 1e-5, 2e-5\}$. During training we sample one negative example from the initial ranked list obtained in Section 4.1, and train the model with pairwise ranking loss.

We use Spacy[1] for sentence splitting, POS tagging and Named Entity Recognition. We use the *en_core_web_lg* model to obtain word vectors.

# 6 RESULTS

In this section we present our experimental results.

## 6.1 Initial retrieval

We examine the performance of the initial retrieval step when different variations of the query $q$ are used. Table 4 shows the results. We observe that, for both datasets, when using both the event $e$ and the context $c$ we get better results than when using either of the two alone, especially in terms of R@1000. This shows that both the event $e$ and the context $c$ are important for our task.

In Table 4 (bottom row) we also show ranking performance when using the link sentence as the query (see Section 3.1). Even though we do not use the link sentence as part of the query in our task definition (Section 2.2), this can give us a reference point for the "upper bound" performance in this step, since the link sentence has a high lexical overlap with the relevant article $d^*$ [34]. We observe that, indeed, when using the link sentence as the query, ranking performance is much higher than when using $q$, achieving close to perfect R@1000. Nevertheless, R@1000 when using $e$ & $c$

[1] http://spacy.io/

Table 5: Retrieval performance when reranking the ranked list obtained by BM25 (first row).

| | WaPo | | Guardian | |
|---|---|---|---|---|
| **Method** | MRR | R@20 | MRR | R@20 |
| BM25 | 0.172 | 0.433 | 0.149 | 0.382 |
| Recency | 0.086 | 0.284 | 0.065 | 0.065 |
| BERT | 0.182 | 0.451 | 0.173 | 0.447 |
| RRF-recency | 0.206 | 0.509 | 0.195 | 0.477 |
| RRF | **0.236** | **0.588** | **0.212** | **0.533** |

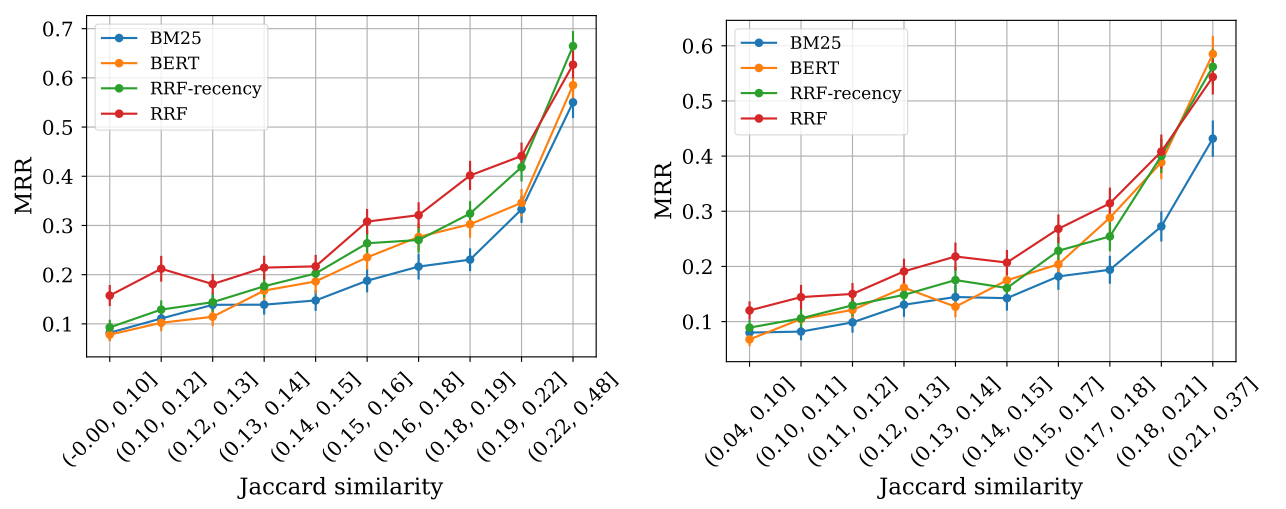

(1) WaPo (2) Guardian

**Figure 3: MRR vs Jaccard similarity between query $q$ and $d^*$.**

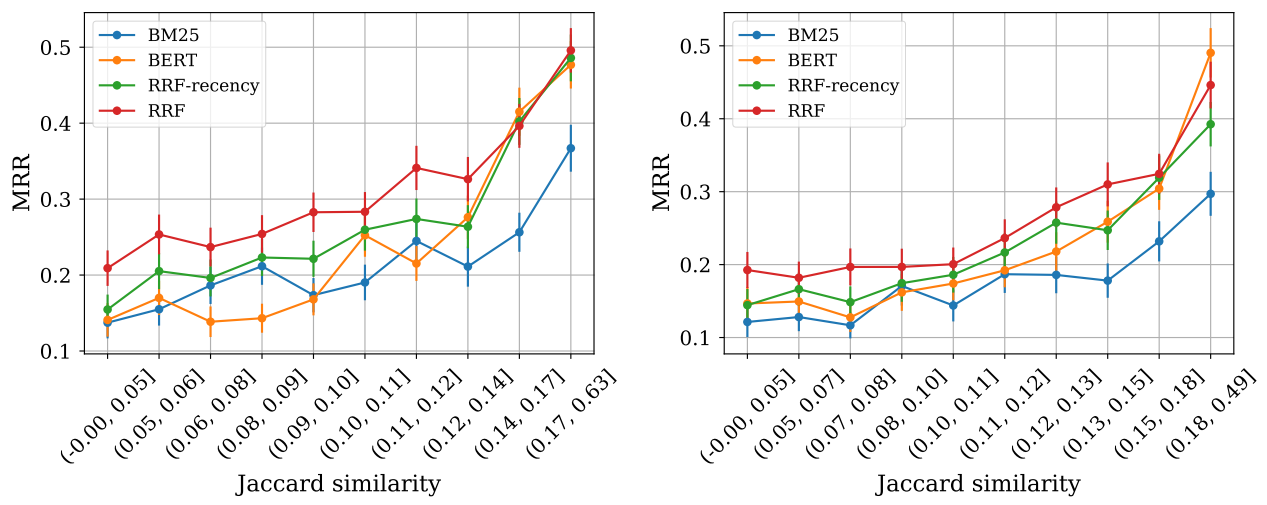

(1) WaPo (2) Guardian

**Figure 4: MRR vs Jaccard similarity between narrative's context $c$ and $d^*$.**

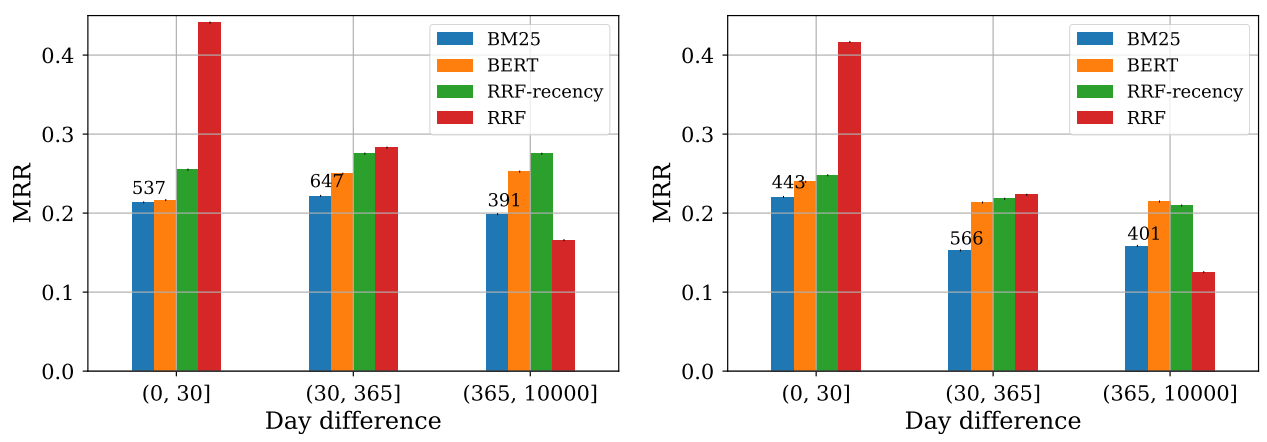

(1) WaPo (2) Guardian

**Figure 5: MRR for retrieval methods grouped per day difference of the query and the relevant article.**

is relatively close to when using LS, which is an encouraging result given that in this step we are more interested in recall.

## 6.2 Reranking

Here, we report results on the individual rankers described in Section 4.2 and their combinations with RRF. Table 5 shows the results. First, we see that the performance of the Recency ranker is poor. Also, we see that BERT outperforms BM25 on both datasets, while only using the headline and the lead of the candidate news article. RRF-recency combines BERT and BM25 achieves an increase over BERT. Finally, when also adding the Recency ranker in RRF, we observe a significant ($p < 0.01$) increase on all metrics. We conclude that RRF, albeit simple, is effective in combining the three rankers and that all three rankers are useful for this task.

# 7 ANALYSIS

In this section we analyze our results along different dimensions to gain further insights into this task. For our analysis we use the development set of the WaPo and Guardian datasets.

## 7.1 Vocabulary gap

The vocabulary gap is a well known challenge in information retrieval [26]. Here, we analyze the performance of the rankers under comparison for this task based on the vocabulary gap between the query $q$ and the relevant article $d^*$, using Jaccard similarity.

In Figure 3 we observe that the higher the lexical overlap between $q$ and $d^*$ (small vocabulary gap) the higher the performance for all rankers, for both datasets. Also, we see that when the lexical overlap is low (large vocabulary gap), all rankers fail to return the relevant article at the top positions of the ranking. This shows that more sophisticated methods are needed to handle the large vocabulary gap in this task. In Figure 4 we show the lexical overlap between the narrative's context $c$ only and the relevant $d^*$. Even though it follows the same trend as in Figure 3, we see that BERT is consistently better than BM25 as the term overlap between the narrative's context $c$ and $d^*$ increases, for both datasets. This shows that BERT is able to better take into account the narrative's context $c$ than BM25.

Next, we show examples of high/low lexical overlap between $q$ and $d^*$ in Table 6. In the first example (high lexical overlap), we see that because of high term overlap, all rankers are able to rank $d^*$ at the top 1–2 positions. In the second example (low lexical overlap), the relevant article $d^*$ discusses the execution of Alfredo Prieto: this is a case in which Morrogh, a prosecutor in Virginia, was involved (Morrogh is mentioned in the narrative's context $c$). However, the fact that Morrogh is involved in the case is not mentioned explicitly in $d^*$ and thus all rankers fail to rank the relevant article at the top positions. Incorporating the fact that Morrogh is related to Prieto in the ranking model could potentially be achieved by exploiting knowledge graphs that store event information [14, 37]. We leave the exploration towards this direction for future work.

## 7.2 Temporal aspects

As discussed in Section 3.2, the retrieval datasets we derived for this task have a strong recency bias. Here, we analyze the performance of the rankers under comparison based on a temporal aspect, i.e., how recent the relevant article is.

In Figure 5 we show the performance of the retrieval methods for different day differences between the query $q$ and the relevant article $d^*$. As expected, we observe that for RRF, which uses the recency signal, the performance increases substantially on average when the relevant article is recent, and decreases when it is older.

Table 6: Examples from the WaPo dev. set with high/low lexical overlap between $q$ and $d^*$ (top/bottom).

| Query $q$ | | Link sentence | Relevant article $d^*$ | | Top-ranked article RRF | | Rank of $d^*$ | | | |
|---|---|---|---|---|---|---|---|---|---|---|
| Query event $e$ | Narrative's context $c$ | | Headline & Lead | Day diff. | Headline & Lead | Day diff. | BM25 | BERT | RRF-recency | RRF |
| What 'arrest' means for the Canadians detained in China — and the epic battle over Huawei . BEIJING — Over the past five months, as Beijing and Washington have exchanged fire on trade and technology, two Canadian men have been held in near-isolation in Chinese detention facilities. | Last week, a Chinese court scheduled Schellenberg's appeal hearing to begin hours after Meng faced an extradition hearing in Vancouver. | After a Canadian court pushed back a decision in Meng's case, the Chinese court announced it would delay a ruling on whether Schellenberg would be put to death. | Chinese court delays ruling on Canadian's death sentence appeal. BEIJING — A Chinese court has delayed ruling on a Canadian man's appeal against his death sentence for drug smuggling, just hours after a Canadian court set a September date for the next hearing in an extradition case against a top Chinese executive. | 7 | $d^*$ | | 1 | 2 | 1 | 1 |
| Fairfax race for prosecutor puts focus on pace of criminal justice reform. Political races are usually about striking contrasts, but in the first Democratic primary for prosecutor in Virginia's largest county in 55 years, both candidates give themselves the same title: progressive. | Morrogh, who was first elected commonwealth's attorney in 2007, has spent nearly all of his career in the prosecutor's office, where he has won some high-profile cases and avoided major scandal. | Morrogh helped secure the convictions of D.C. sniper Lee Boyd Malvo, serial killer Alfred Prieto and more recently the MS-13 gang members who killed a 15-year-old girl. | The execution of Alfredo Prieto: Witnessing a serial killer's final moments. JARRATT, Va. — It is undeniably disturbing to drive to the scheduled killing of another. A hurricane brewing in the distance, slicing steady rain through the gray day. | 1339 | Money from PAC funded by George Soros shakes up prosecutor races in Northern Virginia. A political action committee funded by Democratic megadonor and billionaire George Soros has made large contributions to two upstart progressive candidates attempting to unseat Democratic prosecutors in Northern Virginia primary races. | 38 | 371 | 98 | 151 | 252 |

Table 7: Examples from the WaPo dev. set with a recent relevant article where RRF ranks the relevant article at the top, while RRF-Recency ranks it lower.

| Query $q$ | | Link sentence | Relevant article $d^*$ | | Top-ranked article RRF-recency | | Rank of $d^*$ | | | |
|---|---|---|---|---|---|---|---|---|---|---|
| Query event $e$ | Narrative's context $c$ | | Headline & Lead | Day diff. | Headline & Lead | Day diff. | BM25 | BERT | RRF-recency | RRF |
| China's influence on campus chills free speech in Australia, New Zealand. SYDNEY — Chinese students poured into Australia and New Zealand in the hundreds of thousands over the past 20 years, paying sticker prices for university degrees that made higher education among both countries' top export earners. | After years of feeling fortunate about their economic relationship with China, Australians are starting to worry about the cost. | On Thursday, a ruling party lawmaker, Andrew Hastie, compared China's expansion to the rise of Germany before World War II, suggesting it posed a direct military threat. | Threat from China recalls that of Nazi Germany, Australian lawmaker says. The West's approach to containing China is akin to its failure to prevent Nazi Germany's aggression, an influential Australian lawmaker warned, earning a rebuke from Beijing while highlighting the difficulty the U.S. ally faces in weighing its security needs against economic interests. | 3 | China's meddling in Australia — and what the U.S. should learn from it. While American attention remains focused on Russia's interference in the 2016 presidential election, Australia — perhaps the United States' closest ally — is debating the designs that a different country altogether has on its political system, economy and public opinion. That country is China. | 788 | 32 | 38 | 14 | 1 |
| Despite national security concerns, GOP leader McCarthy blocked bipartisan bid to limit China's role in U.S. transit. House Minority Leader Kevin McCarthy (R-Calif.) blocked a bipartisan attempt to limit Chinese companies from contracting with U.S. transit systems, a move that benefited a Chinese government-backed manufacturer with a plant in his district, according to multiple people familiar with the matter. | Lawmakers frequently take a stance on legislation that could affect campaign contributors or hometown companies. But McCarthy's intervention was striking because the close ally of President Trump sought to protect Chinese interests at a time when Trump and many lawmakers on Capitol Hill are attempting to curb Beijing's access to U.S. markets, particularly in industries deemed vital to national security. | Just last week, Trump put Chinese telecom giant Huawei on a trade "blacklist" that severely restricts its access to U.S. technology. | Trump administration cracks down on giant Chinese tech firm, escalating clash with Beijing. The Trump administration on Wednesday slapped a major Chinese firm with an extreme penalty that makes it very difficult for it to do business with any U.S. company, a dramatic escalation of the economic clash between the two nations. | 5 | Trump says he'll spare Chinese telecom firm ZTE from collapse, defying lawmakers. President Trump said late Friday he had allowed embattled Chinese telecommunications giant ZTE Corp. to remain open despite fierce bipartisan opposition on Capitol Hill, defying lawmakers who have warned that the huge technology company should be severely punished for breaking U.S. law. | 360 | 98 | 9 | 3 | 1 |

Next, we look at specific examples to better understand the results. Table 7 shows examples where the relevant article is recent and RRF ranks it at the top of the ranking, while RRF-recency ranks it lower. In both examples, RRF-recency's top-ranked article seems to also be relevant to $q$, however the writer chose to refer to a more recent event [32]. Note that the fact that only one article is relevant to each query is an artifact of our dataset and not of the task itself.

Table 8 shows examples where the relevant article is old and RRF-recency ranks it at the top of the ranking, while RRF ranks it lower. In the first example, the relevant article discusses a development on the injury of Scherzer, a player of the Washington Nationals team, and RRF-recency correctly brings that at the top position. However, RRF ranks a more recent event at the top position that discusses an injury of a different player of the same team. In the second example, RRF brings at the top position an article that discusses an event about India that is more recent than the one that the relevant article discusses, however the article is off-topic.

The above phenomena suggest that more sophisticated methods that model recency should be explored for this task. For instance, it would be interesting to try to predict which queries are of temporal nature based on the characteristics of the underlying collection [22, 33]. However, methods that build on features derived from user interactions are not applicable to our setting [12].

## 7.3 Entity popularity

Entities play a central role in event-centric narratives, especially in the news domain [37]. We examine whether entity popularity affects retrieval performance in our task by measuring the Inverse Document Frequency (IDF) of entities mentioned in the query [31].

**Table 8: Examples from the WaPo dev. set with an old relevant article where RRF-recency ranks the relevant article at the top, while RRF ranks it lower.**

| Query $q$ | | Link sentence | Relevant article $d^*$ | | Top-ranked article RRF | | Rank of $d^*$ | | | |
|---|---|---|---|---|---|---|---|---|---|---|
| Query event $e$ | Narrative's context $c$ | | Headline & Lead | Day diff. | Headline & Lead | Day diff. | BM25 | BERT | RRF-recency | RRF |
| Max Scherzer's knuckle injury might keep him from being ready for Opening Day. The knuckle at the base of Max Scherzer's right ring finger became the most analyzed joint in the Washington Nationals' clubhouse on Thursday, knocking Stephen Strasburg's right elbow out of its familiar spotlight, and delivering an unexpected blow to the early-season stability of the Nationals' rotation. | Scherzer expected the sprain to heal with regular rest in the offseason. But the symptoms did not improve by December, when another MRI exam revealed the fracture. | A month later, the fracture still had not healed, so he told Team USA Manager Jim Leyland he would not be able to pitch in the World Baseball Classic. | Max Scherzer won't pitch in WBC because of stress fracture in finger. Nationals ace Max Scherzer, one of the first and highest-profile players to commit to play for the United States in the upcoming World Baseball Classic, will not participate in the tournament because of "the ongoing rehabilitation stress fracture in the knuckle of his right ring finger," the club announced Monday afternoon in a statement. | 927 | Nationals place Stephen Strasburg on disabled list (again) with pinched nerve. MIAMI — Nothing appeared amiss for Stephen Strasburg on Wednesday. He played catch at Miller Park in Milwaukee as scheduled, a day before he was to take the mound for the Washington Nationals against the Miami Marlins on Thursday night. But the throwing session didn't go well. | 363 | 12 | 1 | 1 | 3 |
| A mysterious sickness has killed nearly 100 children in India. Could litchi fruit be the cause? NEW DELHI — The children go to sleep as best they can in the sweltering heat. Early in the morning, the fever spikes and the seizures begin. | In August 2017, India witnessed a notorious outbreak of encephalitis in the city of Gorakhpur in the neighboring state of Uttar Pradesh. | More than 30 children died over two days at one hospital after its oxygen ran out. | 'It's a massacre': At least 30 children died in Indian hospital after oxygen is cut off. NEW DELHI — One by one, the infants and children slipped away Thursday night, their parents watching helplessly as oxygen supplies at the government hospital ran dangerously low. | 674 | 'It is horrid': India roasts under heat wave with temperatures above 120 degrees. NEW DELHI — When the temperature topped 120 degrees (49 Celsius), residents of the northern Indian city of Churu stopped going outside and authorities started hosing down the baking streets with water. | 11 | 1 | 1 | 1 | 3 |

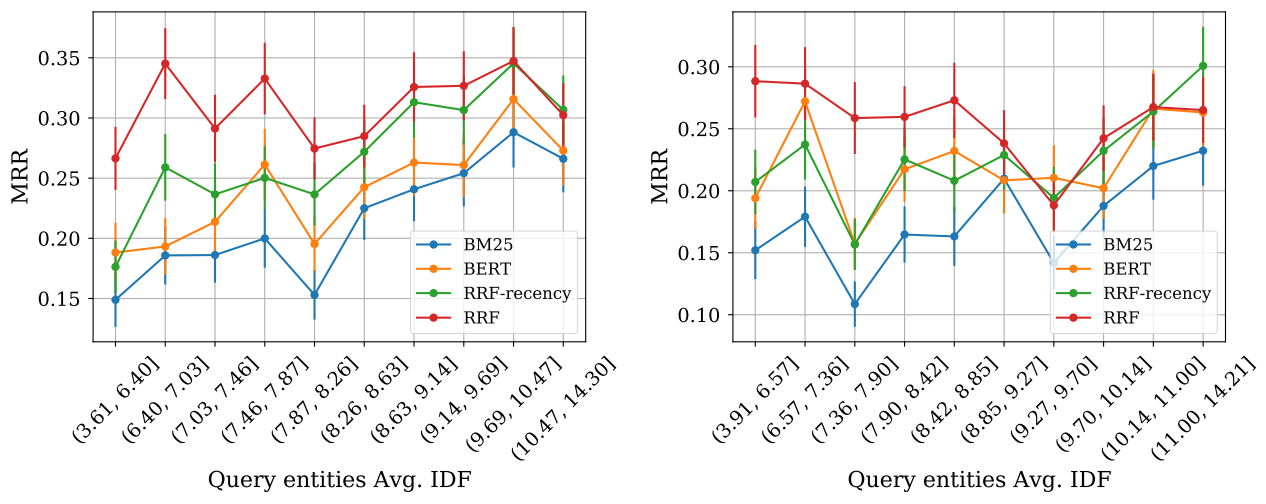

**(1) WaPo** **(2) Guardian**

**Figure 6: MRR vs avg. IDF of the entities in the query $q$.**

An entity with a high IDF in the collection is less popular than an entity with a low IDF.

In Figure 6 we show the performance depending on the average IDF of the entities in the query in the underlying collection. We observe that the rankers that use the query and article text (BM25, BERT, RRF-recency) perform worse for queries with more popular entities (low IDF) than for queries with less popular entities. This is because popular entities appear in multiple events, and thus there are many potentially relevant articles for a query. We also see that RRF, which takes recency into account, is more robust to entity popularity. This might also be related to the fact that a recent event that involves a popular entity is more likely to be relevant in general than a less recent event that involves the same entity (also see examples in Section 7.2, Table 7).

## 7.4 Link sentence

Recall that we do not use the link sentence as part of the query (see Section 3.1). Thus, our rankers are not aware of its content. However, we found that in some cases the link sentence contains information that is crucial for the connection of the *complete* narrative and the relevant news article. Thus, in such cases, the query event $e$ and the narrative's context $c$ are not sufficient. Table 9 shows examples of such cases. Note that in the first example, the relevant article was not even retrieved in the top-1000 of the initial retrieval step (see Section 4.1). In the second example, the relevant article is ranked very low by all rankers.

One direction for future work would be to detect parts of the link sentence that contain such crucial information and add them to the narrative's context $c$. This could be performed as a manual annotation task or modeled as a prediction task [21].

# 8 RELATED WORK

## 8.1 Supporting narrative creation

Recent work on developing automatic applications to support writers has focused on designing tools that track and filter information from social media to support journalists [11, 53]. Cucchiarelli et al. [8] track the Twitter stream and Wikipedia edits to suggest potentially interesting topics that relate to a new event that a writer can include in their narrative when reporting on the event. In contrast, instead of relying on external sources, we aim to retrieve news articles that describe events from the past that can help the writer expand the incomplete narrative about a specific event.

Perhaps the closest to our task are the works by Maiden and Zachos [30] and MacLaughlin et al. [29]. Maiden and Zachos [30] focus on suggesting articles that would help journalists discover new, creative angles on a current incomplete narrative. The difference with our work is that they aim to suggest creative angles on articles and retrieve articles depending on the angle the writer selects. In addition, they evaluate their system in a living lab scenario, whereas we create static retrieval datasets from historical data and use them to train ranking functions. Evaluating our system in a living lab scenario would be a promising direction for future work.

MacLaughlin et al. [29] retrieve paragraphs that contain quotes from political speeches and conference transcripts, so that writers can use them in their incomplete narratives. Even though their retrieval task definition is similar to ours, our task differs in that our unit of retrieval is a news article from a large news article collection instead of a paragraph from a single document (e.g., a political speech). Moreover, our unit of retrieval (article) is timestamped, which makes the temporal aspect prominent in our task.

**Table 9: Examples from the WaPo dev. set where the link sentence contains crucial information for the connection of the complete narrative and the relevant article.**

| Query $q$ | | Link sentence | Relevant article $d^*$ | | Top-ranked article RRF | | Rank of $d^*$ | | | |
|---|---|---|---|---|---|---|---|---|---|---|
| Query event $e$ | Narrative's context $c$ | | Headline & Lead | Day diff. | Headline & Lead | Day diff. | BM25 | BERT | RRF-recency | RRF |
| Americans are drinking more 'gourmet' coffee. This doesn't mean they're drinking great coffee. The National Coffee Association USA recently dropped its annual survey results, and, as usual, there's a wealth of information to sift through to better understand the state of coffee drinking in America. | According to this year's finding, coffee remains the No. 1 drink: Sixty-three percent of the respondents said they drank a coffee beverage (drip coffee, espresso, latte, cold brew, Unicorn Frappuccino, etc.) the previous day, a click down from 64 percent in 2018. | By the way, the second-most consumed beverage was unflavored bottled water, which might help explain the Great Pacific Garbage Patch. | Plastic within the Great Pacific Garbage Patch is 'increasing exponentially,' scientists find. Seventy-nine thousand tons of plastic debris, in the form of 1.8 trillion pieces, now occupy an area three times the size of France in the Pacific Ocean between California and Hawaii, a scientific team reported on Thursday. | 371 | N/A | N/A | N/A | N/A | N/A | N/A |
| Turkey's elections show the limits of Erdogan's nationalism. Ahead of local elections throughout his country last weekend, Turkish President Recep Tayyip Erdogan resorted to his usual tactics. He cast some of his ruling party's opponents as traitors in league with terrorists. | There's a broader story to be told, as well. | Well before President Trump, Indian Prime Minister Narendra Modi or Hungarian Prime Minister Viktor Orban, Erdogan arrived at the politics of the zeitgeist. | Trump's populism is about creating division, not unity. President Trump begins his third week in office with the worst approval ratings of any new American president since polls began tracking such results. | 786 | Stunning setbacks in Turkey's elections dent Erdogan's aura of invincibility. ISTANBUL — Turkish President Recep Tayyip Erdogan faced the prospect Monday of a stinging electoral defeat in Istanbul, the city whose politics he dominated for a quarter of a century, with vote results showing what appeared to be an opposition victory in the race for the city's mayor. | 1 | 447 | 471 | 487 | 535 |

## 8.2 Context-aware citation recommendation

The task of context-aware citation recommendation is to find articles that are relevant to a specific piece of text a writer has generated [17]. It has mainly been studied in the scientific domain [13, 18, 20, 39], but also in the news domain [25]. The main difference between the aforementioned works and our task is that we aim to retrieve articles to expand existing incomplete narratives instead of finding citations for complete narratives.

## 8.3 Event extraction & retrieval

Events are the starting points of narrative news items. Recent work has focused on extracting and characterizing events from large streams of documents [3] and extracting the most dominant events from news articles [5]. In our work, we assume that a news article is associated with a single main event, which is described by the article's headline and lead paragraph [4]. More related to our task is work focused on retrieving events given a query event [25, 40]. However, this work does not consider additional context in the query as we do, and thus it is not directly comparable to ours.

# 9 CONCLUSION AND FUTURE WORK

In this paper, we have proposed and studied the task of news article retrieval in context for event-centric narrative creation. We have proposed an automatic dataset construction procedure and have shown that it can generate reliable evaluation sets for this task. Using the generated datasets, we have compared lexical and semantic rankers and found that they are insufficient. We have found that combining those rankers with one that ranks articles by their reverse chronological order significantly improves retrieval performance over those rankers alone.

As to broader implications of this work, we believe that this work provides new avenues for information retrieval researchers as it introduces and operationalizes a very rich (and realistic) notion of context. We hope that our study will facilitate further work on supporting writers create narratives in the news domain.

As to future work, our analysis has shown that the vocabulary gap for this task is large, and therefore more advanced methods for semantic matching are needed. This could be achieved by incorporating external knowledge about events (and the entities involved in them) in the retrieval methods [14, 37]. Such knowledge includes relationships between entities [47] and sub-event relations [1, 15]. Moreover, our analysis has shown that the temporal aspect is prominent in this retrieval task. Therefore, future work should aim to find more robust ways to incorporate the temporal aspect in the ranking function [22]. A possible way to achieve that is to identify temporal phenomena such as trending terms or entities in the underlying news article collection [22] or in external sources such as social media [8]. Furthermore, we have found that this task is more challenging when the query event involves entities that appear more frequently in the collection, which we plan to further study in the future.

Another direction for future work is to categorize queries in relation to their discourse function in the narrative [42, 43], for example in relation to their function with respect to the main event of the narrative [4], and develop specialized rankers for each category.

Moreover, we have found that in some cases the link sentence contains crucial information for the connection between the *complete* narrative and the relevant news article. However, since the link sentence is not part of the query according to our dataset construction procedure, the constructed query may miss this key piece of information to capture the connection. In future work we aim to ask experts to manually add such missing information to the narrative context. Since this task can be cumbersome, we aim to cast it as a prediction task [21].

Finally, it is important to note that even though our dataset construction procedure can generate reliable retrieval datasets, the fact that we only have a single relevant article for each query may be limiting as more than one article may be relevant. Thus, some of our findings might be an artifact of that procedure and not the task itself. We plan to overcome this limitation in future work by asking journalists to qualitatively assess the output of different rankers to enrich the automatically constructed datasets with more relevant articles per query [29, 30].

## DATA

To facilitate reproducibility, we share the scripts used to generate the datasets used in this paper at `https://github.com/nickvosk/ictir2021-news-retrieval-in-context`.

## ACKNOWLEDGMENTS

This research was supported by the Netherlands Organisation for Scientific Research (NWO) under project nr CI-14-25. All content represents the opinion of the authors, which is not necessarily shared or endorsed by their respective employers and/or sponsors.